\documentclass[conference]{IEEEtran}
\IEEEoverridecommandlockouts

\usepackage{cite}
\usepackage{amsmath,amssymb,amsfonts}
\usepackage{algorithmic}
\usepackage{graphicx}
\usepackage{textcomp}
\usepackage{xcolor}
\usepackage{hyperref}
\usepackage{multirow}
\begin{document}

\title{Cross-View Cross-Modal Unsupervised Domain Adaptation for Driver Monitoring System}

\author{\IEEEauthorblockN{Aditi Bhalla\IEEEauthorrefmark{1}\IEEEauthorrefmark{2}
                         Christian Hellert \IEEEauthorrefmark{2},
                        Enkelejda Kasneci \IEEEauthorrefmark{1}}
        \IEEEauthorblockA{\IEEEauthorrefmark{1}School of Social Sciences and Technology, Technical University Munich, Germany\\
                            Email: \{aditi.bhalla, enkelejda.kasneci\}@tum.de}
        \IEEEauthorblockA{\IEEEauthorrefmark{2}Aumovio SE, Germany}
}

\maketitle

\begin{abstract}
Driver distraction remains a leading cause of road traffic accidents, contributing to thousands of fatalities annually across the globe. While deep learning-based driver activity recognition methods have shown promise in detecting such distractions, their effectiveness in real-world deployments is hindered by two critical challenges: variations in camera viewpoints (cross-view) and domain shifts such as change in sensor modality or environment. Existing methods typically address either cross-view generalization or unsupervised domain adaptation in isolation, leaving a gap in the robust and scalable deployment of models across diverse vehicle configurations. In this work, we propose a novel two-phase cross-view, cross-modal unsupervised domain adaptation framework that addresses these challenges jointly on real-time driver monitoring data. In the first phase, we learn view-invariant and action-discriminative features within a single modality using contrastive learning on multi-view data. In the second phase, we perform domain adaptation to a new modality using information bottleneck loss without requiring any labeled data from the new domain. We evaluate our approach using state-of-the art video transformers (Video Swin, MViT) and multi modal driver activity dataset called \textit{Drive\&Act}, demonstrating that our joint framework improves top-1 accuracy on RGB video data by almost 50\% compared to a supervised contrastive learning-based cross-view method, and outperforms unsupervised domain adaptation-only methods by up to 5\%, using the same video transformer backbone. 
\end{abstract}

\begin{IEEEkeywords}
Driver activity recognition, unsupervised domain adaptation, representation learning, transfer learning
\end{IEEEkeywords}

\section{Introduction}
Global road safety statistics highlight the urgent need for robust Driver Monitoring Systems (DMS) to enhance road safety by identifying and mitigating two leading causes of traffic accidents: driver distraction and fatigue. According to the World Health Organization, approximately 1.19 million people die globally each year due to road traffic accidents, with tens of millions more suffering serious injuries. A significant contributing factor to the increased likelihood of road accidents is distracted driving\footnote{\url{https://www.who.int/news-room/fact-sheets/detail/road-traffic-injuries}}. In the European Union, reports from 2024 indicate approximately 19,800 road fatalities, with driver distraction contributing to an estimated 5–25\% of all crashes in the region\footnote{\url{https://road-safety.transport.ec.europa.eu/system/files/2024-01/ERSO-TR-Distraction_2023-12-19.pdf}}\cite{ETSC2025_DistractionTag}. With the increased use of smartphones and in-vehicle infotainment systems in recent years, it is assumed that there will be more accidents caused by driver distraction~\cite{li2024domain}. Therefore, both academic researchers and the automotive industry are actively working to develop advanced DMS technologies that can monitor driver behavior in real-time, enabling timely interventions and contributing to a safer driving environment.

Standard driver activity recognition (DAR) systems use deep learning-based methods on driver images, video, or physiological data such as gaze or heart rate to estimate driver distraction. These models typically assume that the training and testing data are drawn from the same distribution, including camera angles and environmental conditions. However, this assumption often breaks down in real-world deployments due to variations in sensor placement, changes in driver behavior, or fluctuations in environmental lighting conditions. These shifts can lead to significant domain discrepancies and substantial performance degradation. To address challenges arising for DAR from domain gaps, two complementary approaches are often employed: Unsupervised Domain Adaptation (UDA) techniques and cross-view generalization methods. Cross-view generalization approaches primarily focus on achieving viewpoint invariance, whereas UDA techniques are concerned with addressing changes in the environment or sensor modality. For cross-view generalization, methods such as 3D driver body pose estimation using RGB videos, feature disentanglement, and contrastive learning using images have been used to separate view-dependent and view-invariant features, aiming to learn viewpoint-invariant representations that generalize across different camera angles~\cite{da2022unsupervised},\cite{bianco2024cross},\cite{martin2023viewpoint}. Whereas, UDA-based approaches aim to mitigate domain shift between training and deployment environments by aligning feature distributions between the source and target domains without requiring labeled data from the target domain~\cite{liu2023structure},\cite{lee2021method}, \cite{li2024domain}. 

While collectively, these methods address key challenges in viewpoint variance and domain mismatch, none of the mentioned works explicitly tackle both viewpoint variability and domain shift simultaneously. A unified framework that integrates cross-view invariant representations with domain-adaptive feature alignment for video data is still largely unexplored. Developing such a framework could significantly improve the robustness of real-world DAR systems against unseen camera viewpoints and new deployment environments, all without the need for additional labeling. In this work, we address the challenges of cross-view action recognition (CVAR) and domain adaptation in DMS by proposing a novel two-phase cross-view, cross-modal unsupervised domain adaptation $\mathrm{(C^2UDA)}$ framework. Our main contributions are as follows:
\begin{itemize}
    \item We introduce a unified framework that simultaneously addresses cross-view generalization and domain shift across sensor modalities or deployment environments without requiring additional labeled data from the target domain.
    \item Our framework incorporates a two-phase training strategy: (1) view-invariant feature pretraining using contrastive learning to align multi-view representations within a single modality, and (2) unsupervised cross-modal/domain adaptation using information bottleneck loss to enhance transferability to new modalities or environments.
    \item We validate our approach on two state-of-the-art video transformers Video swin~\cite{liu2022video} and Multiscale vision transformers(MViT)~\cite{fan2021multiscale} using the publicly available \textit{Drive\&Act} dataset~\cite{drive_and_act_2019_iccv}, which features varied viewpoints and modalities.
\end{itemize}

Our experimentation shows that the proposed $\mathrm{(C^2UDA)}$ framework improves generalization across unseen views and modalities in driver activity recognition. On the \textit{Drive\&Act }dataset, it achieves up to 50\% gain over cross-view method using contrastive learning alone, and outperforms UDA-only baselines by up to 5\%, using the same backbone. We further evaluated our trained model on another driver monitoring dataset Driver Anomaly Detection (DAD)~\cite{kopuklu2020driver} to demonstrate the generalization of our method to environment and modality changes.

\section{Related Works}
\noindent\textbf{Driver Activity Recognition.} Due to the growing need for accurate DAR systems, several approaches and modalities have been explored in the last few years. The most widely adopted methods are vision-based, leveraging in-cabin RGB or infrared cameras to monitor the driver’s posture, gaze direction, head orientation, and hand positioning~\cite{ohn2016looking}. Image-based DAR commonly employs convolutional neural networks (CNNs)~\cite{xing2019driver} for feature extraction and classification tasks, with some methods integrating 2D skeleton-based keypoint data using pose estimation tools such as OpenPose~\cite{cao2019openpose} or MediaPipe~\cite{mediapipe} to model driver pose dynamics more effectively~\cite{vats2022key},\cite{lin2021driver}. In addition to static image analysis, video-based DAR approaches aim to capture temporal behavior by utilizing spatiotemporal models such as 3D CNNs~\cite{9377022}, Recurrent neural networks (RNNs)~\cite{jain2016recurrent}, Long short-term memory networks (LSTMs)~\cite{saleh2017driving}, \cite{behera2018context}, and, more recently, Vision Transformers (ViTs)~\cite{liang2022stargazer}, \cite{peng2022transdarc} and TimeSformers~\cite{wang2023sparse}, which excel at learning long-range temporal dependencies. Hybrid models that combine CNNs with LSTM are also commonly used to capture both spatial and temporal characteristics~\cite{cura2020driver}. 

In addition to RGB and infrared modalities, some systems incorporate depth sensors or thermal imaging to improve robustness under low-light conditions~\cite{bar2012driver}. Moreover, multi modal approaches that fuse data from multiple sources, such as vehicle CAN-bus signals, inertial measurement units (IMUs), and eye-tracking sensors, have demonstrated improved performance, particularly in complex scenarios involving subtle distractions or multitasking behaviors~\cite{zhang2020driver}. Unlike our work, these approaches consider a standard supervised setting, where all data is labeled, and there is no domain shift between the training and evaluation datasets.

\noindent\textbf{View-invariant Activity Recognition.} DAR is a specialized subset of activity recognition, where CVAR poses a significant challenge in generalizing models trained on data with specific camera placements. To address this, recent research has focused on learning view-invariant representations that enable generalization across different viewpoints. These approaches can broadly be categorized into skeleton-based and non-skeleton-based methods. Skeleton-based approaches leverage 2D or 3D body keypoints to abstract away view-dependent appearance, focusing on human pose and motion. For instance,~\cite{li20213d} introduced a cross-view contrastive learning framework for unsupervised 3D skeleton-based action representation learning, whereas~\cite{zhang2023cross} proposed learning viewpoint- and condition-invariant dynamics using 3D skeletal or RGB data. In the DAR context,~\cite{martin2023viewpoint} estimated 3D driver pose from in-cabin videos and mapped it to a canonical view to enable viewpoint-invariant classification.

The non-skeleton or appearance-based methods operate directly on RGB or infrared video without explicit pose estimation, aiming to learn view-invariant features from raw data. \cite{ponbagavathi2024probing}~provides a comprehensive evaluation of foundation models for fine-grained action recognition under view changes, highlighting the importance of temporal fusion and architecture design for cross-view robustness. In the area of DAR, \cite{bianco2024cross}~addressed the camera placement challenge by designing DBMNet, a model that utilizes feature disentanglement and contrastive loss to remove viewpoint-specific information while retaining task-relevant behavioral cues, demonstrating strong performance on cross-camera distracted driver classification tasks. These advancements collectively highlight the importance of explicitly modeling viewpoint variation to ensure reliable and scalable DAR systems suitable for real-world, multi-camera environments. However, they do not address simultaneous generalization across both camera viewpoints and domain shifts (e.g., lighting, sensor modality, or deployment context), limiting their robustness in new modalities or environments.

\begin{figure*}[tb]
    \centering
    \includegraphics[width=0.7\linewidth]{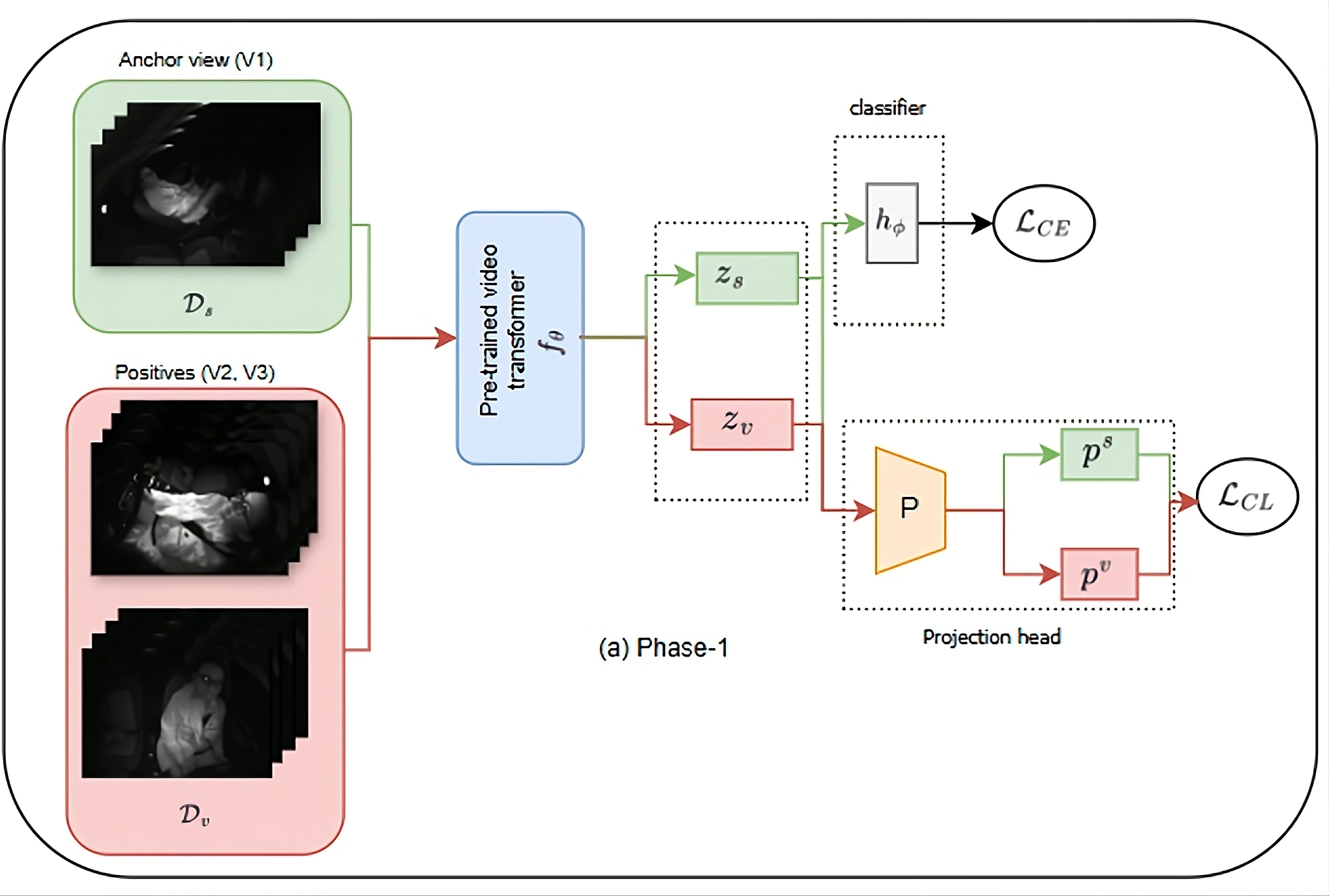} 
    \vspace{1em}
    \includegraphics[width=0.5\linewidth]{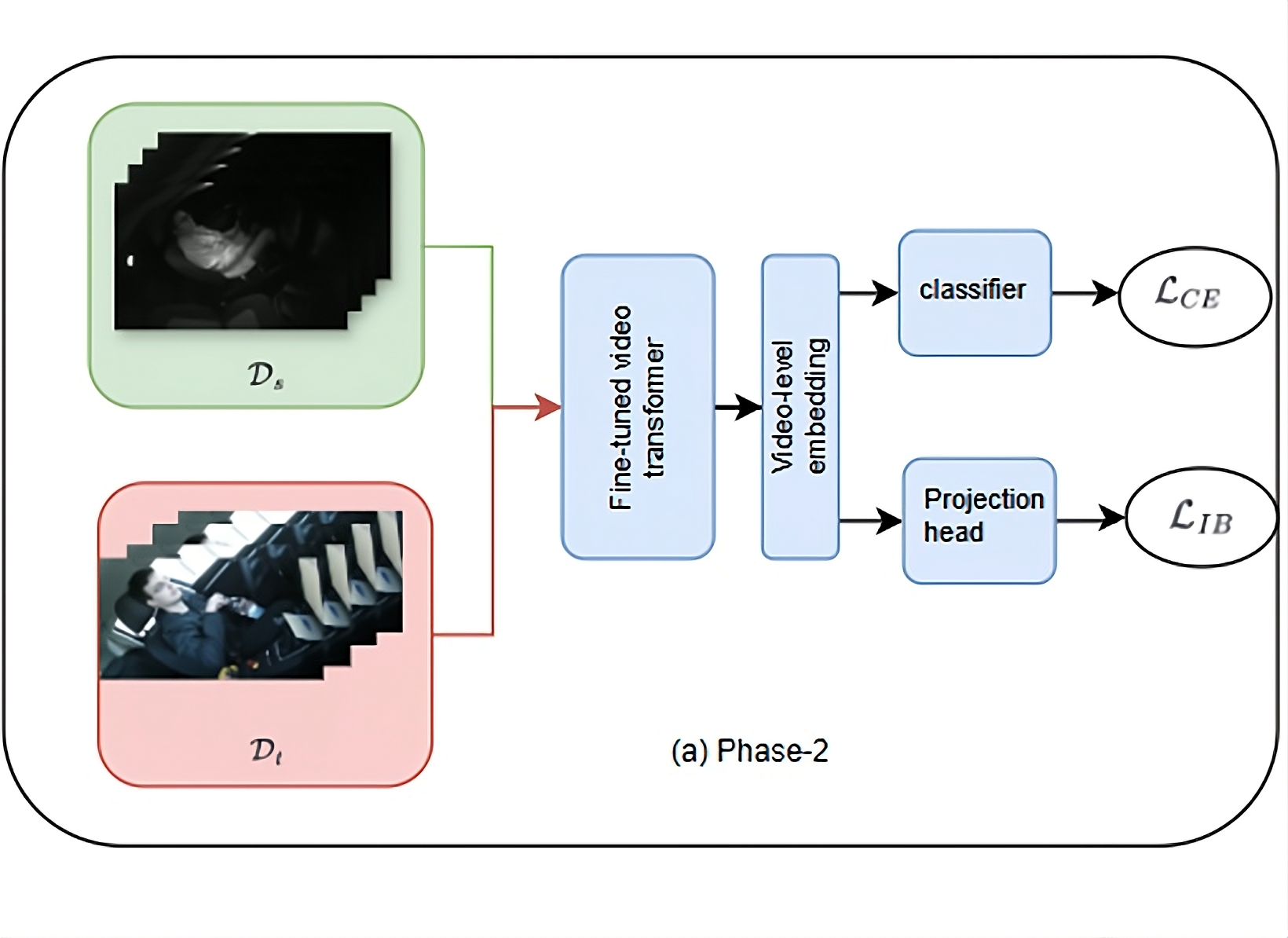}
    \caption{Overview of our $\mathrm{C^2UDA}$ framework. (a) Phase-1: View-invariant pretraining is performed using supervised contrastive learning on synchronized multi-view video data (V1 as anchor, and V2 and V3 as positives) within the same modality (NIR). The shared video transformer backbone learns to produce class-discriminative and view-invariant features via a combined cross-entropy and contrastive loss. (b) Phase-2: The model is adapted to a new modality (RGB) using unlabeled target data. An IB loss is applied between source and target representations to promote domain alignment, while classification supervision is maintained on the source domain.}
    \label{fig:method}
\end{figure*}

\noindent\textbf{Unsupervised Domain Adaptation for Activity Recognition.} Domain adaptation (DA) has emerged as a critical approach to address domain shifts that arise due to varying environments, sensor configurations, or individual behaviors. DA methods are commonly classified based on two key criteria: (a) the availability of labeled data in the target domain- Supervised, semi-supervised, and unsupervised DA, and (b) the relationship between the label space of the source and the target data- closed-set, partial set, and open-set DA~\cite{lee2021method}. UDA assumes no labeled data in the target domain, making it highly applicable to real-world DAR methods, where annotation is costly. Some commonly used approaches to address UDA include discrepancy-based methods, adversarial-based methods, feature-alignment methods, and self-supervised learning-based methods.  

Several works have addressed domain adaptation under diverse conditions for DAR. For instance, \cite{wang2023domain}~proposed a skeleton-guided domain-adversarial network to align feature distributions across domains and~\cite{wang2023driver} introduced a multi-scale domain adaptation network to enhance robustness in distraction detection, whereas~\cite{reiss2020deep} tackled cross-modal adaptation, aligning representations across sensor types using a classification-driven deep learning approach. Beyond DAR, advances in Human Activity Recognition (HAR) literature offer relevant insights, such as hybrid models that separate domain-invariant and domain-specific features~\cite{prabono2021hybrid}, graph-based learning~\cite{yang2024domain}, and sample-weighted adaptation for unlabeled target data~\cite{hu2023swl}. 

More recent works have focused on enhancing structural and temporal robustness in domain adaptation, such as ~\cite{liu2023structure} aligns source and target features while preserving structural information in the activity space,~\cite{lee2021method} adapts to new domains post-deployment through incremental learning, and~\cite{li2024domain} introduces class-wise alignment and confusion regularization to enhance target domain performance.  Particularly relevant is the work by \cite{da2022unsupervised}, who introduce a two-phase UDA framework for video-based action recognition using transformers. In first phase, the spatial and temporal modules are trained on source data, followed by fine-tuning only the temporal module in phase two using an Information Bottleneck(IB)~\cite{tishby2000information}, \cite{tishby2015deep} loss to promote domain invariance. Building on this idea, our work proposes a similar two-phase strategy tailored for cross-view and cross-modal generalization in DAR. Instead of a two-stream transformer architecture that separates spatial and temporal modules, we employ a unified spatiotemporal transformer as the backbone. In Phase 1, we incorporate supervised contrastive learning to enforce view-invariant representation learning within a single modality. In Phase 2, we apply the IB loss to align the source and target feature distributions without requiring labeled data from the target domain. These enhancements not only simplify the architecture but also explicitly integrate view-invariance and domain-adaptation objectives, tackling a significant and previously underexplored gap in the design of robust, real-world DAR systems.

\section{Proposed Method}

\noindent\textbf{Problem setting and notation.} Let $\mathcal{X}$ denotes input space, i.e. video data and $\mathcal{Y}$ represent the set of activity labels. Our goal in this work is to develop a DAR model that generalizes across different camera viewpoints and sensor modalities, such as RGB or near-infrared (NIR), without requiring labeled data from new domains. We define a feature encoder $f_\theta: \mathcal{X} \rightarrow \mathcal{Z}$, which maps an input sample $x \in \mathcal{X}$ to a latent feature representation $z =  f_\theta(x) \in \mathcal{Z}$ and a classifier $h_\phi: \mathcal{Z} \rightarrow \mathcal{Y}$, which maps features to predicted activity labels. Note that $\theta$ and $\phi$ are the learnable parameters. The full model is denoted as: 
\begin{equation}
F_{\theta, \phi}(x) = h_{\phi}(f_{\theta}(x))
\end{equation}
Our setup involves three data domains:
\begin{itemize}
    \item $\mathcal{D}_s= \{(x_i^s,y_i^s)\}_{i=1}^{n_s}$: Source domain with labeled data from one camera view and modality
    \item  $\mathcal{D}_v = \{(x_j^v, y_j^v)\}_{j=1}^{n_v}$: Auxiliary view domain with synchronized labels (inherited from $\mathcal{D}_s$) captured from a different camera viewpoint but the same modality
    \item $\mathcal{D}_t= \{x_k^t\}_{k=1}^{n_t}$: Target domain with unlabeled data from the same view as the source but a different modality or sensor configuration
\end{itemize}

\noindent\textbf{Overview}
An overview of our method $\mathrm{C^2UDA}$ is shown in fig.\ref{fig:method}. We adopt a two-phase training pipeline designed to improve generalization in DAR across both camera viewpoints and sensor modalities. In the first phase, we aim to learn view-invariant representations from the same modality using labeled source data $\mathcal{D}_s$ and auxiliary view data $\mathcal{D}_v$. In the second phase, the model is adapted to a new modality using source data $\mathcal{D}_s$ and unlabeled target data $\mathcal{D}_t$. The key steps of each phase are detailed as follows.

\noindent\textbf{Phase-1: View invariant Pre-training}
The training process in this phase begins with fine-tuning a video transformer encoder $f_\theta$, which is pretrained on a large-scale dataset, as shown in fig.\ref {fig:method}(a). The model consists of a series of spatiotemporal attention blocks that jointly process spatial and temporal information from uniformly sampled video clips $x_i^s \in \mathcal{D}_s$ and $x_j^v \in \mathcal{D}_v$. Each input frame is partitioned into fixed-size patches, linearly projected, and passed through a stack of transformer layers. The spatiotemporal representation is aggregated using a classification token that attends over both spatial and temporal dimensions, producing a video-level embedding $z_s =  f_\theta(x_i^s)$ and $z_v =  f_\theta(x_j^v)$. The source embedding $z_s$ is passed to the classifier $h_{\phi}$, and the model is supervised using a standard cross-entropy loss:
\begin{equation}
   \mathcal{L}_{\text{CE}} = -\frac{1}{n_s} \sum_{i=1}^{n_s} \log p(y_i^s \mid h_\phi(z_s^i)),
\end{equation}

To encourage view-invariant feature learning, both  $z_s$ and $z_v$ are fed into a shared projection head $P$, yielding projected features $p^s = P(z_s)$ and $p^v = P(z_v)$. These projected embeddings are used to compute a contrastive loss $\mathcal{L}_{CL}$, which aligns representations of videos from different viewpoints that belong to the same class. We adopt a supervised contrastive objective \cite{khosla2020supervised}, where all other samples of the same class are treated as positives, while the rest are treated as negatives:
{\small
\begin{equation}
\mathcal{L}_{\text{CL}} = - \frac{1}{2B} \sum_{i=1}^{2B} \frac{1}{|\mathcal{P}(i)|} \sum_{j \in \mathcal{P}(i)} \log \frac{\exp\left(\mathrm{sim}(\mathbf{p}_i, \mathbf{p}_j)/\tau\right)}{\sum_{k=1}^{2B} 1_{[k \neq i]} \exp\left(\mathrm{sim}(\mathbf{p}_i, \mathbf{p}_k)/\tau\right)}
\end{equation}
}

where \( \mathbf{p}_i \) and \( \mathbf{p}_j \) are L2-normalized projected embeddings from the two views, \( \mathrm{sim}(\cdot, \cdot) \) denotes cosine similarity, \( \tau \) is a temperature scaling factor, and \( \mathcal{P}(i) \) is the set of indices sharing the same class label as anchor \( i \). The denominator excludes the anchor index \( i \) to prevent trivial comparisons.

Together, the model is trained using the following objective in Phase 1:
\begin{equation}
    \mathcal{L}_{Phase1} = \mathcal{L}_{CE} + \lambda_1\mathcal{L}_{CL},
\end{equation}
where $\lambda_1$ is a balancing hyperparameter. This phase yields a feature encoder that is not only task-discriminative but also robust to changes in viewpoint, thereby establishing a strong initialization for cross-modal adaptation in Phase 2.

\noindent\textbf{Phase 2: Cross-Modal Adaptation}
In the second phase of our algorithm, we focus on adapting the fine-tuned model to the target domain, leveraging both labeled source data $\mathcal{D}_s$ and unlabeled target data $\mathcal{D}_t$. Unlike \cite{da2022unsupervised}, which decomposes the learning pipeline into separate spatial and temporal modules, we adopt a unified spatiotemporal transformer architecture. This design choice enables the joint modeling of spatial and temporal patterns in video sequences within a single encoder, thereby reducing architectural complexity and enhancing the consistency of learned representations across domains.

In the training phase, we freeze a subset of layers, especially the lower layers responsible for spatiotemporal feature encoding, of the fine-tuned model from phase 1. The remaining parameters, particularly those in the upper layers closer to the classification head, remain trainable. This selective freezing enables us to scale the batch size during adaptation, which is crucial for stable optimization and effective domain alignment.

The adaptation process is shown in fig.\ref{fig:method}(b), which begins by sampling a fixed number of frames from both source and target video clips. The transformer backbone generates video-level embeddings $z_s$ and $z_t =  f_\theta(x_j^t)$ for the source and target domains, respectively. Similar to \cite{da2022unsupervised}, we employ a training objective based on the Information Bottleneck (IB) principle to bridge the domain gap. This framework aims to maximize the mutual information between the learned source features and the corresponding target instances, $I(z_s, D_t)$, while minimizing the mutual information between the features and the original source inputs, $I(z_s,D_s)$, thereby encouraging domain-invariant yet task-relevant representations.

Due to the unavailability of target annotations, we rely on pseudo-labeling strategies to construct positive source-target pairs via a queue-based matching scheme. Unlike augmentation-based methods that rely on intra-instance transformations, our approach pairs each target instance with multiple source samples within the same batch, increasing the number of informative training pairs. This strategy enhances representation robustness and has been empirically shown to improve adaptation performance.

Following prior work, we approximate the IB-based loss using a variant of the Barlow Twins objective. Specifically, we compute a cross-correlation matrix $C$ between normalized feature vectors of source-target pairs and minimize the following IB loss:
\begin{equation}
\mathcal{L}_{\text{IB}} = \sum_{i}^d (1 - C_{ii})^2 + \lambda \sum_{i}^d \sum_{j \ne i}^d (C_{ij})^2
\end{equation}
where $C_{ij}$ measures the correlation between the $i$-th feature dimension of the source and the $j$-th dimension of the target representation. In addition to the IB-based objective, we maintain a classification loss \(\mathcal{L}_{CE}\) computed using the labeled source domain samples. This helps preserve class-discriminative information during the adaptation process. Specifically, the video-level embeddings $z_s$ of source samples are passed through a classifier head $h_{\phi}$ trained with cross-entropy loss. Therefore, the total loss used during phase 2 is a weighted combination of the two objectives:
\[
\mathcal{L}_{\text{total}} = \mathcal{L}_{CE} + \alpha \mathcal{L}_{IB},
\]
where \(\alpha\) is a hyperparameter balancing domain alignment and classification performance.

\section{Experiments}

\noindent\textbf{Dataset.} We evaluate our proposed $\mathrm{C^2UDA}$ framework using the \textit{Drive\&Act} dataset \cite{drive_and_act_2019_iccv}, a large-scale, multi-modal, and multi-view benchmark for driver activity recognition. The dataset features over 12 hours and 9.6 million frames of people engaged in distracting activities, captured in color, depth, and infrared videos, using multiple synchronized cameras placed at five different locations: right-top (V1), left-top (V2), front-top (V3), back (V4), and face-view (V5) in a realistic car cabin environment. This setup enables a detailed investigation of viewpoint variability and modality shifts in a real-world setting.

To construct a controlled experimental setup for Phase-1 (cross-view generalization), we utilize three camera views (V1, V2, and V3) from the NIR modality, capturing driver activity from various top profiles. Using temporal metadata and class-wise organization of \textit{Drive\&Act} we apply a custom multi-view synchronization procedure to align video clips from the three views that correspond to the same activity class and temporal window. These synchronized clips serve as semantically aligned samples across views. For our supervised contrastive learning setup, we treat V1 as the anchor and use the other two views (V2 and V3) as positive examples, leveraging the fact that all three views share the same class label. This encourages the model to learn view-invariant, class-discriminative representations by contrasting samples with the same label across different viewpoints. We used a 70-15-15 split strategy to divide the data into training, validation, and test sets, ensuring balanced class distribution and sufficient samples per view.

For Phase-2 (cross-modal domain adaptation), we treat the NIR domain as the labeled source and the RGB domain as the unlabeled target. Both modalities share the same viewpoint (V1), but differ significantly in appearance due to lighting and sensor characteristics, resulting in a natural domain shift. This setup enables us to evaluate the model's capacity to transfer learned knowledge from NIR to RGB without requiring labeled data in the target modality.

\noindent\textbf{Implementation details.} Our method was implemented using PyTorch~\cite{paszke2019pytorch} and related Python libraries. In Phase 1, we initialized the backbone video transformer model pre-trained on Kinetics-400~\cite{kay2017kinetics} provided by PyTorch\footnote{\url{https://pytorch.org/vision/main/models/video_swin_transformer.html}, \url{https://pytorch.org/vision/main/models/video_mvit.html}}. Fine-tuning was performed for 20 epochs using the Adam optimizer, a cosine learning rate decay schedule, an initial learning rate of 0.001, weight decay of $10^{-9}$, and a batch size of 8 video clips per domain. To promote domain-general representation learning during this phase, all model layers were made trainable.

In Phase 2, we performed partial fine-tuning to adapt the model to the unlabeled target domain. Specifically, the lower spatiotemporal encoding layers were frozen, while the upper layers and classifier head were fine-tuned. Training was conducted for an additional 20 epochs using the same optimizer, learning rate scheduler, and weight decay. However, the learning rate was increased to 0.005, and the batch size was raised to 64 video clips per domain to support stable domain alignment under the IB loss, following the approach described in~\cite{da2022unsupervised}. This design facilitates robust feature alignment between source and target domains during unsupervised adaptation.

\noindent\textbf{Baselines.} To assess the effectiveness of our proposed method, we compare it against standard baseline approaches using video transformer backbones. In particular, we evaluate on multiple input views (V1 to V4) and backbones (Video Swin and MViT), reporting Top-1 and Top-5 accuracy. The first baseline method involves directly fine-tuning the backbone pretrained on Kinetics-400 using labeled data from the V1 view. The second baseline extends this by incorporating a view-level contrastive objective between the source view (V1) and auxiliary views (V2 and V3), encouraging alignment of representations across views while still relying on supervised fine-tuning with labeled V1 data. The third baseline adopts a two-phase UDA strategy with IB loss, but without using contrastive loss during model fine-tuning in first phase. Finally, our method, which performs  cross-modal UDA using IB loss, and incorporates supervised contrastive learning in the first phase to fine-tune the spatiotemporal transformer backbone. To ensure fairness, all baseline models were trained under the same optimization schedule. 

\begin{table*}[t]
\caption{Performance Comparison of Video Transformer Backbones Across Views and Modalities Using Fine-Tuning, Contrastive Learning, and Unsupervised Domain Adaptation}
\centering
\renewcommand{\arraystretch}{1.5}  

\begin{tabular}{|p{2.5cm}|c|cc|cc|cc|cc|cc|}
\hline
\multirow{2}{*}{\textbf{Method}} & \multirow{2}{*}{\textbf{Backbone}} 
& \multicolumn{2}{c|}{\textbf{NIR (V1)}} 
& \multicolumn{2}{c|}{\textbf{NIR (V2)}} 
& \multicolumn{2}{c|}{\textbf{NIR (V3)}} 
& \multicolumn{2}{c|}{\textbf{NIR (V4)}}
& \multicolumn{2}{c|}{\textbf{RGB (V1)}} \\
\cline{3-12}
& & \rule{0pt}{2.5ex} Top-1 & Top-5 & Top-1 & Top-5 & Top-1 & Top-5 & Top-1 & Top-5 & Top-1 & Top-5 \\
\hline
\multirow{2}{*}
{\parbox[c][7ex][c]{2cm}{\centering\shortstack{\textbf{Fine-tuning}\\\ensuremath{V1}}}} 
& \rule{0pt}{2.5ex} VideoSwin & 87.54\% & 97.88\% &15.48\% & 53.18\% & 24.73\% & 57.37\% & 11.41\% & 36.10\%  & 35.96\% & 76.67\% \\
\cline{2-12}
& \rule{0pt}{2.5ex} MViT      & 88.95\% & 98.14\% & 22.29\% & 57.74\% & 34.58\% & 66.97\% & 21.71\% & 56.06\%  & 41.99\% &82.69\% \\
\hline
\multirow{2}{*}{%
\parbox[c][7ex][c]{2cm}{\centering \shortstack{\textbf{Fine-tuning w/} \\ \textbf{Contrastive loss} \\ \ensuremath{V1,V2,V3 \leftrightarrow V4}}}}

& \rule{0pt}{3ex} VideoSwin & 87.28\% & 98.33\% & 83.40\% & 96.04\% & 78.39\% & 91.73\% & 33.91\% & 74.57\%  & 34.56\% &78.78\%\\
\cline{2-12}
& \rule{0pt}{3ex} MViT      & 89.98\% &98.45\% & 84.22\% & 96.89\% & 80.88\% & 93.01\% & 38.63\% & 79.39\%  & 37.11\%  & 82.48\%\\
\hline
\multirow{2}{*}{%
  \parbox[c][7ex][c]{2cm}{\centering \shortstack{\textbf{UDA}\\\ensuremath{NIR \leftrightarrow RGB}}}
}
& \rule{0pt}{3ex} VideoSwin & 87.28\% & 98.33\% & 18.89\% & 54.34\% & 22.34\% & 53.33\% & 13.91\% & 37.64\%  & 86.62\% & 98.81\%\\
\cline{2-12}
& \rule{0pt}{3ex} MViT      & 89.98\% &98.45\% & 21.17\% & 58.15\% & 35.22\% & 64.83\% & 21.62\% & 57.25\%  & 88.68\% &98.16\%\\
\hline
\multirow{2}{*}{%
  \parbox[c][7ex][c]{2cm}{\centering \shortstack{\textbf{Our method}\\\ensuremath{V1,V2,V3 \leftrightarrow V4}\\\ensuremath{NIR \leftrightarrow RGB}}}
}
& \rule{0pt}{3ex} VideoSwin & 88.37\% & 99.10\% & 82.72\% & 97.23\% & 83.27\% & 94.55\% & 32.35\% & 76.22\%  & 87.42\% &97.98\% \\
\cline{2-12}
& \rule{0pt}{3ex} MViT      & \textbf{89.42\%} & \textbf{99.17\%} & \textbf{84.91\%} & \textbf{97.47\%} & \textbf{85.16\% }& \textbf{97.83\% }& \textbf{38.92\%} & \textbf{80.24\%}  & \textbf{89.25\% } & \textbf{98.94\%} \\
\hline
\end{tabular}
\label{tab:results}
\end{table*}

\section{Results}

\begin{table}[t]
    \caption{Perfromance evaluation of models trained using Drive\& Act dataset on DAD Dataset}
    \centering
    \renewcommand{\arraystretch}{1.5} 
    \begin{tabular}{|p{2.7cm}|c|c|c|c|}
    \hline
         \multirow{2}{*}{\textbf{Method}} 
         & \multicolumn{2}{c|}{\textbf{Video Swin}} & \multicolumn{2}{c|}{\textbf{MViT}} \\
         \cline{2-5}
         & \rule{0pt}{2.5ex} Top-1 & Top-5 & Top-1 & Top-5  \\
         \hline
         \textbf{Fine-tuning} & 7.87\% & 30.23\% & 9.89\% & 32.45\%  \\
         \hline
          \parbox[c][7ex][c]{2cm}{\shortstack{\textbf{Fine-tuning w/} \\ \textbf{Contrastive loss}}} & 8.23\%& 31.34\%& 9.74\% & 33.95\%  \\
         \hline
 \textbf{UDA} & 14.59\% & 41.54\% & 16.93\% & 43.87\% \\
         \hline
         \textbf{Our method} & \textbf{21.58\%} & \textbf{55.87\%} & \textbf{24.94\%} & \textbf{57.87\%}  \\
         \hline
    \end{tabular}

    \label{tab:resultDAD}
\end{table}

We perform a comprehensive evaluation of our proposed method against established baselines using two state-of-the-art video transformer backbones, Video Swin and MViT. Table \ref{tab:results} reports Top-1 and Top-5 accuracy across four input views (V1–V4) and two modalities (NIR and RGB). The results highlight the advantages of integrating supervised contrastive learning and cross-modal domain adaptation.

Our method consistently outperforms all baseline approaches across views and modalities. Notably, using MViT, our approach achieves a Top-1 accuracy of 89.25\% on RGB data, which represents a substantial improvement over contrastive-only (37.11\%) and UDA-only (88.68\%) baselines. Across NIR data, our model achieves strong performance on V2 and V3 views, which were used as auxiliary positives during supervised contrastive learning. More importantly, it generalizes well to V4, an unseen view during training, achieving up to 38.92\% Top-1 accuracy on MViT. This demonstrates the model’s ability to learn robust, view-invariant representations that transfer effectively to novel viewpoints. These findings validate the benefit of combining view-level contrastive learning with cross-modal unsupervised domain adaptation to address both spatial and modality-level distribution shifts.

To further evaluate generalization beyond the training domain, we conduct a cross-dataset evaluation using the \textit{DAD} dataset, which includes infrared (IR) video data from top view. Table~\ref{tab:resultDAD} presents the performance of models trained solely on the \textit{Drive\&Act} dataset and tested on \textit{DAD} without any fine-tuning.

The results show that our method generalizes substantially better to this unseen domain compared to the baselines. Using MViT, our approach achieves 24.94\% Top-1 and 57.87\% Top-5 accuracy, outperforming the UDA only baseline. Even Video Swin shows notable gains, with our method achieving 21.58\% Top-1, compared to only 7.87\% for standard fine-tuning and 14.59\% for UDA. These improvements demonstrate our method’s robustness to domain shifts not only in modality and view, but also across datasets with different recording conditions, sensor configurations, and subject distributions.

\section{Discussion} These findings have significant practical implications for real-world driver monitoring systems, where sensor placement and lighting conditions vary substantially between different vehicle models and environments. The significant improvements over contrastive baselines demonstrate that leveraging multiple auxiliary views through supervised contrastive learning enables the model to learn view-invariant representations. Similarly, improvements over UDA-only baselines confirm that contrastive pretraining provides stronger feature representations for effective downstream adaptation. 

However, our method has certain limitations. The approach relies on synchronized multi-view data during training, which may not always be available in all deployment environments. Additionally, while we demonstrate cross-dataset generalization to the DAD dataset, domain shifts involving drastically different subjects, camera setups, or activity labels may require further fine-tuning or adaptation. Even with these limitations, the findings are still significant as they confirm that structured representation learning across different views and modalities greatly enhances robustness.

\section{Conclusions}
In this work, we presented $C^2UDA$, a novel unified framework for cross-view and cross-modal unsupervised domain adaptation in driver monitoring systems, leveraging state-of-the-art video transformer models. Unlike existing methods that address viewpoint variation or domain shift in isolation, our approach jointly tackles both challenges through a two-phase training strategy: view-invariant pretraining using supervised contrastive learning, followed by cross-modal adaptation using an information bottleneck objective. Our extensive experiments on the \textit{Drive\&Act} dataset demonstrate that $C^2UDA$ significantly improves generalization across both unseen viewpoints and sensor modalities. Moreover, our cross-dataset evaluation on the Driver Anomaly Detection (DAD) dataset confirms the model’s robustness to domain shifts in real-world settings. Therefore highlighting its practical utility in diverse vehicle configurations without requiring labeled data in the target domain. These findings suggest that unified representation learning across various views and modalities is a promising approach for developing scalable, generalizable, and deployable driver activity recognition systems, particularly in automotive applications where sensor configurations and environments vary.

\textbf{Acknowledgment:}
This work is a result of the joint research project STADT:up (19A22006F). The project is supported by the German Federal Ministry for Economic Affairs and Energy (BMWE), based on a decision of the German Bundestag. The author is solely responsible for the content of this publication.

\bibliography{IEEEabrv,mybibfile}
\end{document}